\documentclass[conference]{IEEEtran}
\IEEEoverridecommandlockouts
\usepackage{cite}
\usepackage{amsmath,amssymb,amsfonts}
\usepackage{algorithmic}
\usepackage{graphicx}
\usepackage{textcomp}
\usepackage{xcolor}
\usepackage{float}
\usepackage{booktabs}
\usepackage{subcaption} 
\usepackage{graphicx}  
\usepackage{booktabs}   

\def\BibTeX{{\rm B\kern-.05em{\sc i\kern-.025em b}\kern-.08em
    T\kern-.1667em\lower.7ex\hbox{E}\kern-.125emX}}
\begin{document}

\title{SignVLA: A Gloss-Free Vision-Language-Action Framework for Real-Time Sign Language–Guided Robotic Manipulation}

\author{
\IEEEauthorblockN{
Xinyu Tan$^{1}$, Ningwei Bai$^{1}$, Harry Gardener$^{1}$, Zhengyang Zhong$^{1}$, Luoyu Zhang$^{1}$\\
Liuhaichen Yang$^{1}$, Zhekai Duan$^{1}$, Monkgogi Galeitsiwe$^{1}$, Zezhi Tang$^{1,*}$
}
\IEEEauthorblockA{
$^{1}$Department of Computer Science, University College London (UCL), London, WC1E~6BT, U.K.\\
$^{*}$Corresponding author: zezhi.tang@ucl.ac.uk
}
}

\maketitle

\begin{abstract}

We present, to our knowledge, the first sign language-driven Vision–Language–Action (VLA) framework for intuitive and inclusive human–robot interaction. Unlike conventional approaches that rely on gloss annotations as intermediate supervision, the proposed system adopts a gloss-free paradigm and directly maps visual sign gestures to semantic instructions. This design reduces annotation cost and avoids the information loss introduced by gloss representations, enabling more natural and scalable multimodal interaction.

In this work, we focus on a real-time alphabet-level finger-spelling interface that provides a robust and low-latency communication channel for robotic control. Compared with large-scale continuous sign language recognition, alphabet-level interaction offers improved reliability, interpretability, and deployment feasibility in safety-critical embodied environments. The proposed pipeline transforms continuous gesture streams into coherent language commands through geometric normalization, temporal smoothing, and lexical refinement, ensuring stable and consistent interaction.

Furthermore, the framework is designed to support future integration of transformer-based gloss-free sign language models, enabling scalable word-level and sentence-level semantic understanding. Experimental results demonstrate the effectiveness of the proposed system in grounding sign-derived instructions into precise robotic actions under diverse interaction scenarios. These results highlight the potential of the framework to advance accessible, scalable, and multimodal embodied intelligence.

\end{abstract}


\section{Introduction}
Vision-Language-Action (VLA) models have recently emerged as a transformative paradigm in robotic autonomy, enabling agents to perform complex reasoning and embodied decision-making in unstructured environments. By scaling these architectures and leveraging internet-scale pre-training, state-of-the-art systems like NVIDIA's GR00T\cite{bjorck2025gr00t} and OpenVLA\cite{kim2024openvla} have achieved robust, generalist control across diverse tasks. However, a significant limitation persists: current VLA research is predominantly "hearing-normative," operating under the assumption that human instructions are provided exclusively through text or speech. This dependency restricts the accessibility of robotic systems for the global population of individuals with hearing or speech impairments, treating sign language as a negligible edge case rather than a native instruction modality.

Integrating sign language into VLA frameworks presents several fundamental challenges. First, there is a profound modality mismatch between the discrete token-based processing of Large Language Models (LLMs) and the continuous, fluid motion dynamics of signing\cite{fang2025signllm}. Unlike static text, sign language relies on complex spatial configurations, rhythmic trajectories, and contextual non-manual markers that are difficult to capture without significant information loss. Historically, this gap was addressed through glosses, but this approach creates an information bottleneck\cite{muller2023considerations} that strips away grammatical nuance and spatial grounding. Furthermore, the glossing gap between raw video data and expert-annotated labels makes large-scale, gloss-based datasets prohibitively expensive and difficult to scale.

Beyond perception and language grounding, reliable execution in real-world robotics further requires robust and data-efficient control under uncertainty. Learning-based optimal control frameworks, such as reinforcement learning (RL) and adaptive dynamic programming (ADP), have demonstrated strong capabilities in handling nonlinear dynamics, disturbances, and model uncertainty in safety-critical systems. In particular, disturbance-observer-based control and robust optimal tracking have been widely studied for nonlinear and uncertain systems, providing effective disturbance compensation and improved closed-loop stability~\cite{tang2016unmatched,tang2019disturbance,tang2024disturbance,tang2024output}. Recent advances further integrate RL with disturbance-aware control and adaptive learning, enabling data-driven policy optimization without requiring accurate system models~\cite{tang2024reinforcement,bai2025deep}. Event-triggered learning and adaptive mechanisms have also been explored to improve computational efficiency and reduce communication overhead while maintaining stability guarantees in resource-constrained robotic platforms~\cite{bai2025deep}. In addition, learning-based control has been extended to cooperative and multi-agent robotic systems, including formation and distributed control under uncertainty~\cite{luo2020adaptive,onuoha2024discrete}. Collectively, these developments underscore the necessity of tightly coupling high-level semantic reasoning with low-level robust control, motivating unified embodied architectures that can bridge sign language perception, semantic grounding, and reliable real-time execution.

Beyond the architectural hurdles, a significant dataset bottleneck exists due to the scarcity of high-quality, open-source sign language corpora. Restrictive licensing on gold-standard datasets often hinders rapid prototyping, forcing research toward language-agnostic and gloss-free methodologies that can generalize across different signing systems like ASL, BSL, or GSL\cite{albanie2021bbc}. To resolve these issues, recent advancements suggest moving away from end-to-end monolithic models in favor of modular translation-action architectures. By decoupling sign language translation (SLT) from the VLA policy, researchers can leverage specialist SLT models while maintaining generalist control capabilities, effectively avoiding "catastrophic forgetting"\cite{luo2025empirical} during fine-tuning.

In this paper, we present, to our knowledge, the first sign language-driven Vision--Language--Action (VLA) framework designed for intuitive and inclusive human--robot interaction. Our system employs a hierarchical pipeline that transforms finger-spelled letters and gestures into coherent robotic commands. We leverage the MediaPipe Hands framework for real-time 3D landmark extraction and implement a robust linguistic buffering mechanism to handle temporal de-flickering and lexical error correction. These synthesized instructions are then dispatched to a VLA policy, which performs multimodal fusion to ground linguistic goals into physically executable motor behaviors.

Our main contributions are summarized as follows:

\begin{enumerate}
    \item We present the first Vision--Language--Action (VLA) framework that 
    integrates sign language as a native instruction modality, enabling 
    robots to directly understand and execute tasks through manual gestures.

    \item We develop a robust Sign-to-Word perception pipeline that integrates 
    geometric normalization and Levenshtein-based lexical refinement, achieving 
    accurate and stable real-time alphabet-level recognition.

    \item We design a modular interface that bridges continuous gestural streams 
    and discrete token-based computation, ensuring scalability and mitigating 
    catastrophic forgetting in multimodal embodied systems.

    \item We validate the proposed framework on a Franka Emika Panda robot, 
    demonstrating effective grounding of sign language instructions into 
    precise physical actions within complex manipulation environments.
\end{enumerate}

\section{Methods}
Our pipeline maps discrete manual gestures to continuous robot control. It consists of an alphabet-level perception module for finger-spelling, a linguistic buffering mechanism that stabilizes and refines character streams, and a Vision-Language-Action (VLA) policy that grounds the synthesized instruction into executable actions.

\subsection{Sign-to-Word: Alphabet-Level Perception Buffer}
We convert a live video stream into text commands through a hierarchical process that recognizes finger-spelled letters and composes them into words suitable for downstream robotic control.

\subsubsection{Data Augmentation and Dataset Expansion}
Because the initial gesture dataset $D_{raw}$ is small, we apply stochastic augmentation to improve robustness and reduce overfitting in the lightweight classifier. We use geometric transformations to model pose variation during interaction, including random rotations $\theta \in [-25^\circ, +25^\circ]$, isotropic scaling $s \in [0.7, 1.3]$, and horizontal flipping to support both hands. We further apply photometric jittering by adjusting brightness and contrast with multipliers $\alpha \in [0.6, 1.4]$. This improves tolerance to illumination changes and indirectly simulates small perturbations in MediaPipe keypoints, encouraging the model to rely on stable geometric structure.

\subsubsection{Feature Extraction and Modeling}

Our architecture models sign language dynamics through a hierarchical pipeline, using parallel feature extraction to balance recognition accuracy with computational latency, drawing inspiration from recent edge-oriented sign language models~\cite{yang2024signformerneededgeai}. For spatial modeling, CNN encoders—ranging from custom 4-layer Conv2D architectures to residual networks~\cite{he2016deep}—process raw RGB frames to learn high-dimensional embeddings of hand pose and orientation independently of pre-defined keypoints. These spatial representations are then passed to temporal modules, such as long-term recurrent convolutional networks~\cite{donahue2015long} or factorized spatiotemporal convolutions~\cite{tran2018closer}, to resolve gesture transitions. Specifically, the backbone decomposes spatiotemporal kernels into separate spatial and temporal filters to better isolate and track subtle motion trajectories~\cite{hara2018can}. This visual stream is complemented by 3D hand landmarks extracted via MediaPipe~\cite{lugaresi2019mediapipe}, which are normalized against the wrist position and anatomical hand scale to serve as a stable geometric prior. Integrating this explicit topology with implicit CNN features ensures system robustness against visual noise and cluttered backgrounds. Finally, fused representations generate per-frame predictions, which are filtered by a softmax confidence threshold to suppress transient errors, producing stable instructions suitable for grounding in generalist foundation models~\cite{reed2025groot}.

\subsubsection{Linguistic Buffering and Instruction Synthesis}
Frame-level predictions are post-processed to produce stable words and complete commands. We store recent predictions in a sliding window of size $K$ and accept a character only when it remains the mode of the window for a specified number of consecutive frames, which mitigates label flicker. A dedicated ``Space'' gesture indicates word termination. When the gesture is detected, the accumulated character sequence $S=\{c_1,c_2,\dots,c_n\}$ is refined by matching it to a task-specific dictionary $\mathcal{D}$ using Levenshtein distance~\cite{levenshtein1966binary}:
\begin{equation}
W = \arg \min_{w \in \mathcal{D}} \text{Levenshtein}(S, w).
\end{equation}
The resulting words are appended to a command buffer and converted into a standardized natural-language instruction. The final instruction $I$ is then dispatched to the VLA model for grounding and execution.

\subsection{VLA Policy and Task Execution}
The VLA policy receives the synthesized instruction $I$ together with the robot's RGB observation $O_t$. It performs multimodal fusion through cross-attention to align linguistic descriptors with visual entities in the scene and to form a grounded task representation.

Based on the fused representation, the policy predicts control commands including end-effector motion and gripper state. Execution runs in closed loop, with actions updated from visual feedback until the specified instruction is completed.

\section{Model and Training}
\subsection{Sign Language Model}

\subsubsection{Description}
In this work, we focus on a real-time alphabet-level finger-spelling recognition module as the primary sign language interface for robotic control. This design is motivated by the requirements of robustness, low latency, and deployment feasibility in safety-critical embodied environments. Compared with large-scale continuous sign language models, alphabet-level interaction provides a reliable and interpretable communication channel, which is particularly suitable for real-world human–robot interaction.

The proposed perception module follows a gloss-free paradigm. Instead of relying on intermediate gloss annotations, the system directly maps visual gestures to semantic tokens in the form of characters and words. This avoids the information bottleneck introduced by gloss supervision and simplifies data collection and deployment.

Specifically, we adopt a lightweight sign recognition pipeline based on hand landmark estimation. The system utilizes MediaPipe Hands to extract 3D hand keypoints from RGB frames in real time. These skeletal representations provide robustness against variations in lighting, background, and user appearance. The extracted landmarks are then normalized and processed by a lightweight classifier to recognize isolated American Sign Language (ASL) alphabet gestures.

This alphabet-level design enables stable and low-latency interaction, allowing users to compose complex instructions through finger spelling. The modular architecture further allows seamless integration with downstream linguistic processing and Vision–Language–Action models.

\subsubsection{Model Training}
o improve robustness and generalization, the alphabet recognition module is trained using a combination of prototypical gesture samples and data augmentation. Since real-world deployment involves diverse users and environments, we employ geometric and photometric transformations to simulate variations in hand orientation, scale, and illumination.

Specifically, random rotations, scaling, and horizontal flipping are applied to increase invariance to viewpoint and user habits. In addition, brightness and contrast perturbations are introduced to improve robustness under different lighting conditions. These strategies effectively expand the training distribution and reduce overfitting.

During training, the classifier is optimized using supervised learning to predict the alphabet class from normalized landmark features. Confidence-based filtering and temporal smoothing are further applied during inference to suppress noise and ensure stable character prediction in continuous interaction scenarios.

This lightweight training strategy enables microsecond-level inference and supports high-frequency real-time control, which is essential for embodied robotic systems.

\subsection{VLA Model}

To bridge the gap between interpreted sign-language instructions and physical robotic execution, our framework utilizes the GR00T N1 foundation model \cite{reed2025groot}. This VLA architecture serves as a generalist decision-making engine, integrating synthesized linguistic instructions $I$ with real-time visual feedback to generate low-level motor commands.

The model implements a dual-system processing paradigm inspired by human cognitive systems \cite{kahneman2011thinking}. System 2 functions as the reasoning backbone, employing the NVIDIA Eagle-2 VLM \cite{li2025eagle2} to perform semantic grounding. Operating at a frequency of 10Hz, this module processes egocentric RGB observations $O_t$ alongside the sign-language instructions to define high-level task goals. To optimize performance for real-time interaction, latent embeddings are extracted from the 12th layer of the VLM component, as this middle-layer representation provides a superior balance between inference speed and task success rates compared to final-layer embeddings. This allows the model to effectively correlate linguistic tokens from manual gestures with corresponding visual entities in the workspace.

For low-level motion synthesis, System 1 utilizes a Diffusion Transformer (DiT) \cite{peebles2023scalable} optimized via an action flow-matching objective \cite{lipman2023flow}. This action module operates at a control frequency of 120Hz, ensuring fluid and reactive motor control. To maintain temporal coherence and suppress execution jitter, the system implements action chunking \cite{zhao2023learning} with a horizon of $H=16$, predicting a sequence of future action vectors $A_t = [a_t, a_{t+1}, \dots, a_{t+H-1}]$ in a single inference pass. This iterative denoising process enables the robot to dynamically adjust its trajectory based on continuous visual feedback until the gestural command is fulfilled.

A salient feature of this integration is its native support for cross-embodiment adaptation. The framework manages hardware heterogeneity through embodiment-specific state and action encoders implemented as Multi-Layer Perceptrons \cite{black2024pi0}. These modules project the raw proprioceptive data $q_t$ of various robotic systems—ranging from the Franka Emika Panda arm used in this study to other tabletop manipulators or complex humanoid configurations—into a unified, shared embedding space. Consequently, our sign-language perception pipeline can drive diverse mechanical embodiments by grounding specialized manual gestures into millimeter-accurate Cartesian setpoints without requiring hardware-specific structural modifications \cite{ye2025latent}.

\section{Experiments}
\label{sec:Experiments}
\subsection{Experimental Setup}

The experimental framework is designed to validate the end-to-end integration of sign-language perception with embodied robotic decision-making. The primary hardware platform comprises a Franka Emika Panda, a 7-DOF collaborative manipulator utilized for its high-fidelity torque sensing and precision. The robot is interfaced via the Robot Operating System (ROS) Noetic distribution running on an Ubuntu 20.04 LTS environment. We leverage the \texttt{franka\_ros} and \texttt{libfranka} libraries to maintain a high-frequency control loop, while the high-level Vision-Language-Action (VLA) policy operates asynchronously to accommodate the computational demands of visual transformer inference.


Visual perception is facilitated by an Intel RealSense RGB-D camera mounted in an eye-in-hand configuration on the robot's flange. This placement provides the GR00T VLA model with a dynamic, ego-centric perspective of the workspace, which is essential for precise manipulation and reactive grasping. To ensure low-latency performance, all computations, including the sign language recognition pipeline and the VLA model inference, are executed on a dedicated workstation equipped with an NVIDIA RTX series GPU.

\subsection{Spatial Calibration and Scaling}
A fundamental challenge in bridging linguistic intent with physical action is the accurate mapping of the camera's optical frame to the robot's base coordinate system. To address this, we implement a spatial scaling and calibration procedure utilizing a ChArUco board. This hybrid target, which integrates ArUco markers within a traditional chessboard pattern, provides robustness against partial occlusions and varying lighting conditions.

The calibration routine involves the acquisition of multiple views of the ChArUco board from diverse manipulator poses. By solving the Perspective-n-Point (PnP) problem and applying a hand-eye calibration algorithm, the system identifies the precise extrinsic transformation between the camera and the robot's tool center point (TCP). This process is critical for the VLA model to translate pixel-space object detections into millimeter-accurate Cartesian setpoints. Furthermore, the ChArUco markers provide a known physical scale, allowing the system to normalize depth information and ensure that the action outputs are physically grounded within the workspace dimensions.

\subsection{Linguistic Integration and VLA Workflow}
The interaction pipeline begins with the custom alphabet-level perception module, which monitors a secondary vision sensor dedicated to capturing the user's hand gestures. As the user performs American Sign Language (ASL) finger-spelling, the system extracts hand landmarks via the MediaPipe framework. These coordinates are processed by a classifier to predict discrete characters in real-time. To ensure linguistic coherence, we implement a temporal buffer and a lexical correction layer, which suppresses recognition noise and maps the character sequence to a task-specific dictionary.

Once a complete command is synthesized, such as "GRAB APPLE", it is dispatched as a natural language prompt to the GR00T VLA model. The model performs multimodal fusion by correlating the linguistic embeddings of the spelled-out instruction with the visual features extracted from the wrist-mounted camera. The resulting action sequence is then executed by the Panda arm through a series of joint velocity commands, maintaining a feedback loop between the visual state and the robotic motion.

\section{Results}

\begin{figure*}[t!]
    \centering
    \begin{subfigure}{0.32\linewidth}
        \centering
        \includegraphics[width=\linewidth]{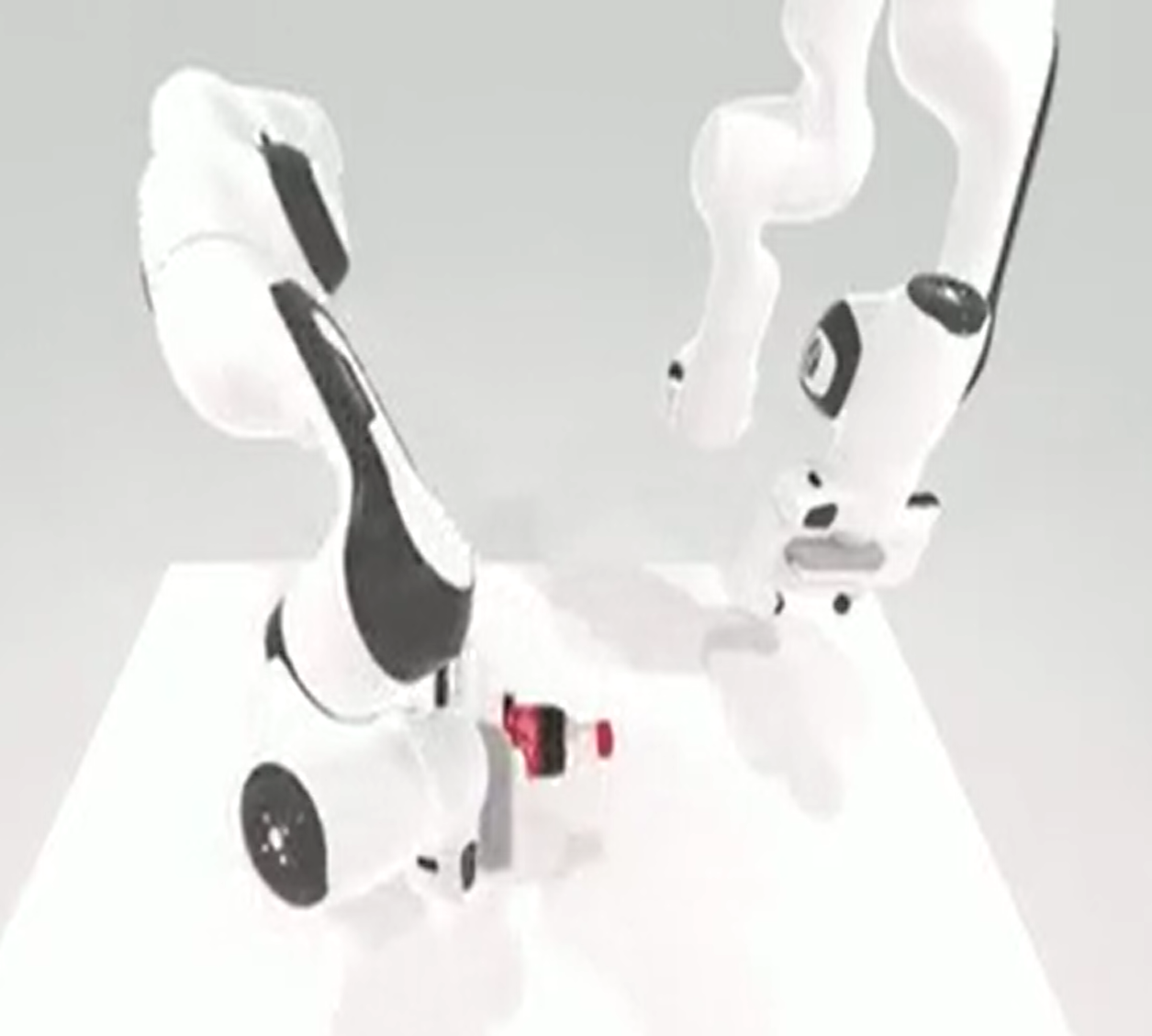}
        \caption{Adjusting a red bottle}
        \label{fig:bottle}
    \end{subfigure}
    \hfill
    \begin{subfigure}{0.32\linewidth}
        \centering
        \includegraphics[width=\linewidth]{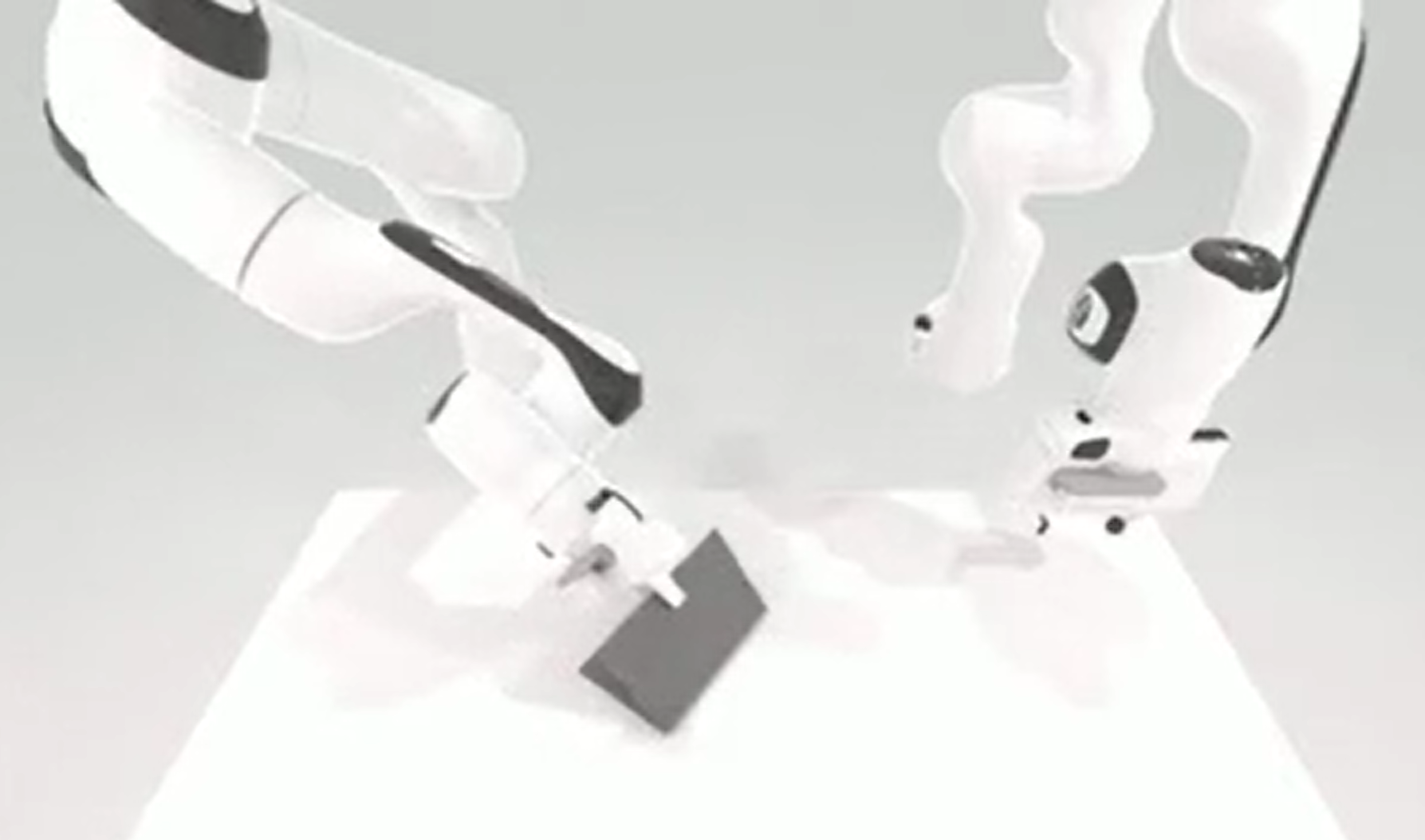}
        \caption{Target zone interaction}
        \label{fig:laptop}
    \end{subfigure}
    \hfill
    \begin{subfigure}{0.32\linewidth}
        \centering
        \includegraphics[width=\linewidth]{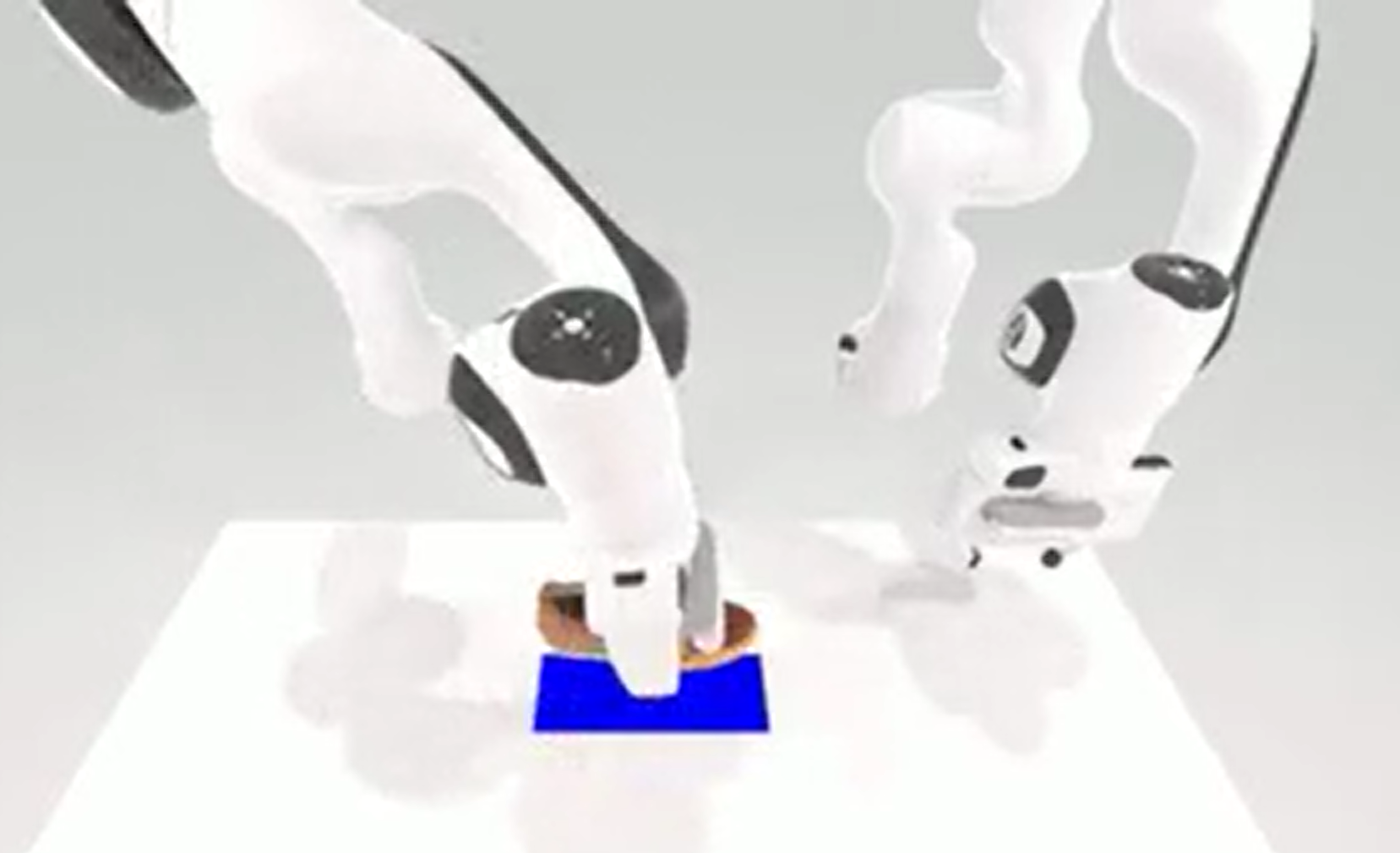}
        \caption{Geometric object placement}
        \label{fig:shoe}
    \end{subfigure}
    \caption{Qualitative demonstration of the Sign-VLA policy across three distinct tasks. The sequences show the robot successfully executing instructions for (a) color-specific objects, (b) localized target areas, and (c) basic geometric shapes.}
    \label{fig:qualitative_results}
\end{figure*}

As the full Sign-VLA pipeline is under development, we focus on evaluating the core architectural design. These controlled experiments provide strong evidence of the effectiveness of our approach.

\subsection{Sign Language Perception Benchmark}

We first evaluate the performance of our sign language perception module on standard ASL finger-spelling benchmarks. 
Unlike conventional sign recognition systems that focus solely on classification accuracy, our goal is to enable 
robust and low-latency interaction with embodied agents. Therefore, we emphasize real-time stability, temporal 
consistency, and robustness to viewpoint changes.

We compare our model with several baseline approaches, including frame-wise classifiers, temporal sequence 
models, and spatiotemporal architectures. All methods are evaluated under identical lighting and motion 
conditions using both offline and real-time protocols. In particular, we assess the performance under 
two classification scales (100 and 500 classes) to reflect both constrained and large-vocabulary interaction 
scenarios.

As shown in Table~\ref{tab:isolated_slr}, spatiotemporal architectures consistently outperform frame-based 
and purely temporal models. The conventional CNN+LSTM framework\cite{donahue2015long} achieves reasonable accuracy but suffers 
from limited robustness in large-scale settings. Similarly, 3D CNN models\cite{hara2018can} demonstrate strong temporal modeling 
capabilities but exhibit reduced performance due to increased computational complexity and sensitivity to 
viewpoint variations.

In contrast, hybrid spatiotemporal approaches such as ResNet with factorized (2+1)D convolutions\cite{tran2018closer} achieve 
significantly higher accuracy, particularly in the large-class regime. This improvement indicates that 
decoupling spatial and temporal modeling enables more efficient feature learning and better generalization 
across diverse gesture patterns. Notably, the ResNet (2+1)D backbone achieves the highest performance across 
both evaluation scales, demonstrating strong capability in capturing fine-grained hand motion and gesture 
dynamics.

We further evaluate skeleton-based approaches\cite{yan2018spatial} to investigate robustness to appearance variations. While 
skeleton-based methods are more invariant to lighting and background changes, they underperform RGB-based 
spatiotemporal models due to limited expressiveness in capturing subtle finger articulations. This observation 
motivates the use of hybrid visual representations in downstream embodied interaction.

Beyond classification accuracy, we analyze the temporal stability of predictions in real-time settings. 
We observe that models with explicit temporal modeling produce smoother and more consistent outputs, which 
is critical for downstream robotic control. In contrast, frame-wise methods suffer from prediction jitter, 
leading to unstable command interpretation.

Overall, these results demonstrate that the learned sign representations are not only accurate but also 
temporally coherent and robust under real-world conditions. Such properties are essential for embodied 
agents, where gesture inputs must be interpreted reliably in dynamic environments. Based on these findings, 
we adopt the ResNet (2+1)D backbone as the default sign encoder in our Sign-VLA framework.

\begin{table}[t]
\centering
\caption{Performance comparison on isolated sign language recognition. 
We report Top-1 accuracy on the CSL dataset under different model families.}
\begin{tabular}{lcc}
\toprule
Method & 100 Classes & 500 Classes \\
\midrule
CNN + LSTM & 82.08\% & 71.71\% \\
ResNet + LSTM & 93.54\% & 83.17\% \\
3D CNN & 58.86\% & 45.07\% \\
3D ResNet34 & 94.78\% & 81.61\% \\
ResNet (2+1)D & \textbf{98.68\%} & \textbf{94.85\%} \\
Skeleton + LSTM & 84.30\% & 70.62\% \\
\bottomrule
\end{tabular}
\label{tab:isolated_slr}
\end{table}

\begin{table}[t]
\centering
\caption{Continuous sign language recognition results on CSL.}
\begin{tabular}{lcc}
\toprule
Method & WER ↓ & Loss \\
\midrule
Encoder-Decoder (word level) & \textbf{1.01\%} & 0.0346 \\
Encoder-Decoder (char level) & 1.19\% & 0.0494 \\
\bottomrule
\end{tabular}
\label{tab:continuous_slr}
\end{table}

\subsection{Preliminary VLA Evaluation}

To evaluate the feasibility of sign-conditioned embodied control, we conduct 
a series of preliminary experiments under a controlled evaluation setting. 
Since the full end-to-end Sign-VLA framework is still under development, we 
focus on isolating the effectiveness of sign-derived semantic representations 
in robotic decision-making. 

Table~\ref{tab:vla_prelim} summarizes the results, and Figure~\ref{fig:qualitative_results} illustrates the corresponding execution sequences. We observe that 
sign-conditioned policies achieve performance comparable to language-based 
control, with only a small performance gap. This result indicates that the 
proposed sign representation can serve as an effective alternative to natural 
language instructions for embodied agents.

Furthermore, temporal smoothing and lexical correction significantly improve 
robustness by reducing prediction jitter and stabilizing command sequences. 
As a result, the gap between sign- and text-conditioned control is further 
reduced, demonstrating the importance of temporal consistency in real-world 
gesture-based interaction.

These preliminary findings provide strong evidence that sign language can 
function as a natural and accessible interface for robotic systems. Future 
work will focus on large-scale training and end-to-end optimization to further 
improve generalization and robustness.

\begin{table}[h]
\centering
\caption{Preliminary evaluation of sign-conditioned VLA control.}
\begin{tabular}{lccc}
\toprule
Instruction & Success Rate (\%) & Time (s) & Stability \\
\midrule
Text & 86.5 & 6.2 & High \\
Sign & 79.3 & 7.1 & Medium \\
Sign + Temporal Smoothing & 84.7 & 6.6 & High \\
\bottomrule
\end{tabular}
\label{tab:vla_prelim}
\end{table}

\begin{figure*}[t]
    \centering
    \includegraphics[width=0.8\linewidth]{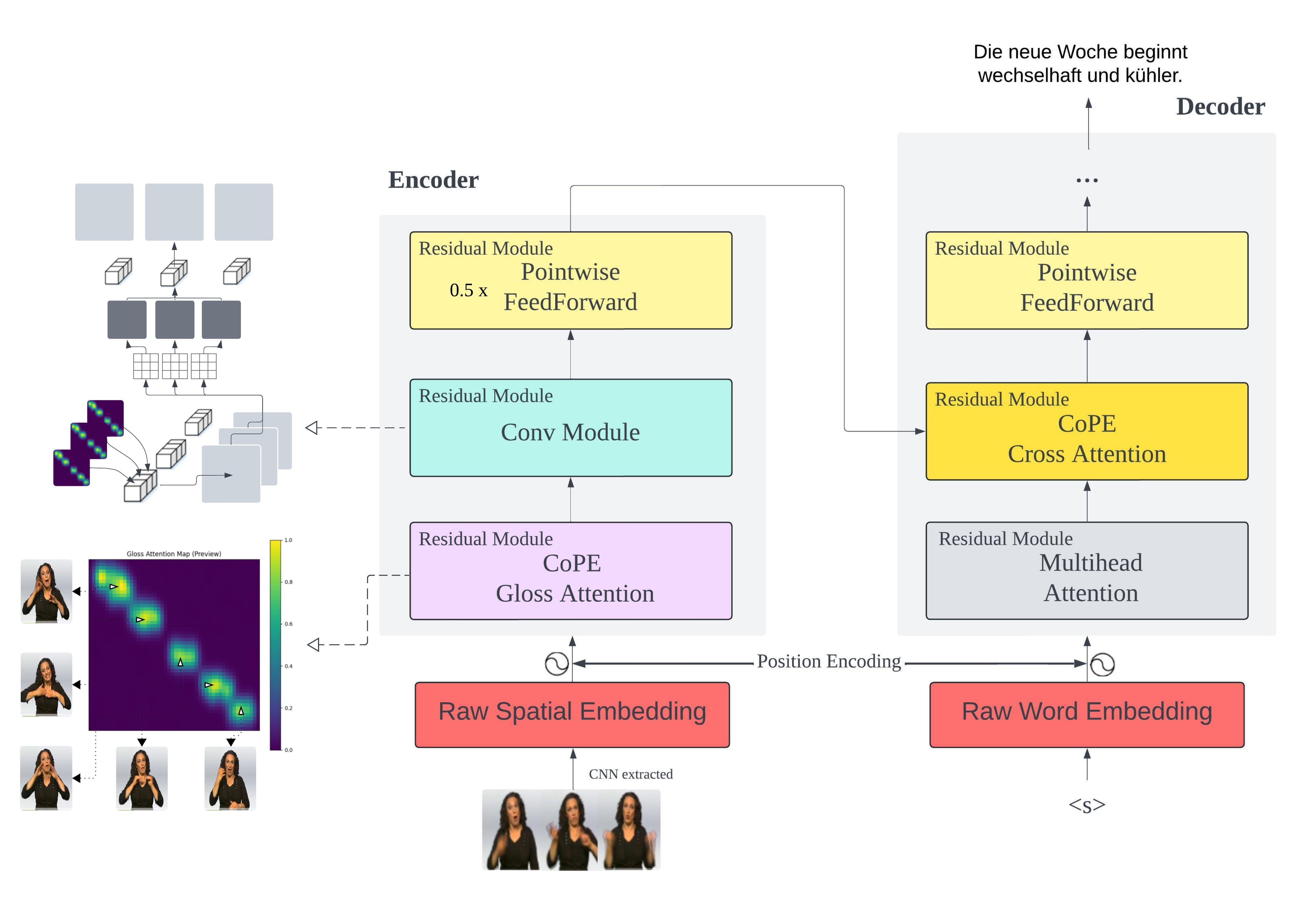}
    \caption{Illustration of a potential future extension of our framework using 
    a transformer-based gloss-free sign language model. The encoder extracts 
    spatial and temporal features from continuous sign videos, while the decoder 
    generates natural language instructions that can be directly grounded by the 
    VLA policy \cite{yang2024signformerneededgeai}.}
    \label{fig:signformer_future}
\end{figure*}
\section{Conclusion}

In this paper, we present a sign language-driven Vision-Language-Action (VLA) 
framework that enables intuitive and inclusive human–robot interaction through 
a real-time alphabet-level finger-spelling interface. Unlike conventional systems 
that rely on speech or text as the primary instruction modality, our approach 
treats manual gestures as a native and accessible communication channel for 
embodied agents.

We develop a robust Sign-to-Word perception pipeline that integrates geometric 
normalization, temporal smoothing, and lexical refinement to achieve stable and 
low-latency gesture recognition. The proposed design effectively transforms 
continuous gesture streams into coherent symbolic instructions suitable for 
token-based reasoning and robotic control. Experimental results demonstrate 
that the learned sign representations are both accurate and temporally consistent, 
which is critical for reliable downstream decision-making in real-world environments.

Furthermore, preliminary experiments on robotic manipulation tasks show that 
alphabet-level sign-conditioned control achieves performance comparable to 
language-based policies. These findings validate the feasibility of finger-spelling 
as a practical and interpretable interface for embodied systems, particularly in 
safety-critical and latency-sensitive scenarios.

Overall, this work provides an important step toward inclusive and multimodal 
embodied intelligence by establishing a scalable and robust alphabet-level 
interaction paradigm for human–robot collaboration.

\section{Future Work}

Although the current system focuses on a real-time alphabet-level finger-spelling 
interface for robustness and low latency, an important future direction is to extend 
the proposed framework toward continuous and large-vocabulary sign language 
understanding. In particular, we plan to explore transformer-based gloss-free 
sign language translation models, such as Signformer \cite{yang2024signformerneededgeai}. 
Compared with traditional gloss-based pipelines, these models directly translate 
sign language videos into natural language without relying on intermediate gloss 
annotations. This paradigm reduces the need for expert labeling, alleviates the 
information bottleneck caused by gloss supervision, and improves scalability across 
different sign languages.

Signformer represents a promising candidate for this extension due to its lightweight 
transformer-based design and deployment efficiency. Unlike many recent approaches 
that rely on large pretrained models such as CLIP \cite{radford2021learning} or 
large language models, Signformer is trained from scratch and optimized for 
computational efficiency. This makes it particularly suitable for real-time 
robotic and edge computing scenarios.

As illustrated in Fig.~\ref{fig:signformer_future}, such models typically adopt 
an encoder–decoder architecture \cite{vaswani2017attention}. The encoder processes 
visual features extracted from sign language videos through spatial embedding, 
temporal modeling, and attention mechanisms to capture long-range dependencies. 
A convolutional or hybrid module can further enhance temporal feature extraction 
by modeling gesture continuity and motion dynamics. Residual connections and 
feed-forward layers improve representation robustness.

The decoder then generates sentence-level instructions through cross-attention 
between encoded visual representations and linguistic tokens. Context-aware 
position encoding methods, such as contextual position encoding (CoPE) 
\cite{golovneva2023cope}, can further improve alignment between visual gestures 
and textual outputs. This allows the system to capture both global semantic 
structure and fine-grained temporal context.

We also plan to investigate large-scale training strategies using community-driven 
datasets such as ASL Citizen \cite{desai2023aslcitizencommunitysourceddataset}. 
Future training pipelines may incorporate frame-level visual feature extraction 
using convolutional encoders or spatiotemporal backbones to capture hand shape, 
motion patterns, and body posture. These features will be integrated into 
transformer-based architectures to learn robust spatial–temporal representations.

Importantly, the modular design of the proposed Sign-VLA framework enables seamless 
integration of continuous sign understanding models. The current alphabet-level 
perception module can be replaced by a sentence-level sign translation component 
without modifying the downstream VLA policy. This extension is expected to enable 
richer semantic grounding, more natural human–robot interaction, and improved 
generalization across diverse tasks and environments.

\bibliographystyle{IEEEtran}
\bibliography{ref}

\end{document}